\documentclass{article}

\usepackage[numbers]{natbib}
\usepackage[preprint]{neurips_2022}

\usepackage[dvipsnames]{xcolor}         
\definecolor{linkColor}{rgb}{0.18,0.39,0.62}
\usepackage[utf8]{inputenc} 
\usepackage[T1]{fontenc}    
\usepackage[colorlinks=true,linkcolor=linkColor,citecolor=linkColor,filecolor=linkColor,urlcolor=linkColor]{hyperref}       
\usepackage{url}            
\usepackage{booktabs}       
\usepackage{amsfonts}       
\usepackage{nicefrac}       
\usepackage{microtype}      
\usepackage{tcolorbox}

\usepackage{graphicx}
\usepackage{arydshln}
\usepackage{booktabs}
\usepackage{multirow}
\usepackage{caption}
\usepackage{subcaption}
\usepackage{makecell}
\usepackage{csquotes}
\usepackage{epigraph}

\RequirePackage{algorithm}
\RequirePackage{algorithmic}

\usepackage{multirow}
\usepackage{amsmath}
\usepackage{capt-of}
\usepackage{tabularx}
\usepackage{epsfig}
\usepackage{amssymb}
\usepackage{amsfonts}
\usepackage{booktabs}
\usepackage{scalerel}
\usepackage[inline]{enumitem}
\usepackage{listings}
\usepackage{varwidth}
\usepackage[export]{adjustbox}
\usepackage{tikz}
\usetikzlibrary{tikzmark}

\usepackage{stmaryrd}
\usepackage{bbm}
\usepackage{wrapfig}
\usepackage{pifont}
\usepackage[noabbrev]{cleveref}

\definecolor{deepblue}{rgb}{0,0,0.5}
\definecolor{officeblue}{RGB}{0,102,204}
\definecolor{deepred}{rgb}{0.6,0,0}
\definecolor{deepgreen}{rgb}{0,0.5,0}
\definecolor{mybrickred}{RGB}{182,50,28}

\definecolor{fillcolor}{RGB}{216,217,252}

\usepackage{etoolbox}
\usepackage{framed}

\newif\ifxetexorluatex
\ifxetex
  \xetexorluatextrue
\else
  \ifluatex
    \xetexorluatextrue
  \else
    \xetexorluatexfalse
  \fi
\fi
\newcommand*\quotesize{60} 
\newcommand*{\openquote}
   {\tikz[remember picture,overlay,xshift=-4ex,yshift=-2.5ex]
   \node (OQ) {\fontsize{\quotesize}{\quotesize}\selectfont``};\kern0pt}

\newcommand*{\closequote}[1]
  {\tikz[remember picture,overlay,xshift=4ex,yshift={#1}]
   \node (CQ) {\fontsize{\quotesize}{\quotesize}\selectfont''};}

\colorlet{shadecolor}{white}

\newcommand*\shadedauthorformat{\emph} 

\newcommand*\authoralign[1]{%
  \if#1l
    \def\authorfill{}\def\quotefill{\hfill}
  \else
    \if#1r
      \def\authorfill{\hfill}\def\quotefill{}
    \else
      \if#1c
        \gdef\authorfill{\hfill}\def\quotefill{\hfill}
      \else\typeout{Invalid option}
      \fi
    \fi
  \fi}
{\authoralign{#1}
\ifblank{#2}
   {\def\shadequoteauthor{}\def\yshift{-2ex}\def\quotefill{\hfill}}
   {\def\shadequoteauthor{\par\authorfill\shadedauthorformat{#2}}\def\yshift{2ex}}
\begin{snugshade}\begin{quote}\openquote}
{\shadequoteauthor\quotefill\closequote{\yshift}\end{quote}\end{snugshade}}

\usepackage{amsmath,amsfonts,bm}

\def\eqref#1{equation~\ref{#1}}

\def\1{\bm{1}}

\DeclareMathAlphabet{\mathsfit}{\encodingdefault}{\sfdefault}{m}{sl}
\SetMathAlphabet{\mathsfit}{bold}{\encodingdefault}{\sfdefault}{bx}{n}

\newcommand\our{\textsc{LongViT}}
\newcommand\longnet{\textsc{LongNet}}

\usepackage{pifont}

\definecolor{bluecode}{RGB}{0, 150, 199}

\title{When an Image is Worth 1,024 × 1,024 Words: \\ A Case Study in Computational Pathology}

\author{
  Wenhui Wang$^\dag$,~~Shuming Ma$^\dag$,~~Hanwen Xu$^{\dag\ddag}$,~~Naoto Usuyama$^\dag$,~~Jiayu Ding$^\dag$ \\
  \textbf{Hoifung Poon$^\dag$,~~Furu Wei$^\dag$} \\
  $^\dag$Microsoft Research \\
  $^\ddag$Paul G. Allen School of Computer Science and Engineering, University of Washington, WA \\
  {\href{https://aka.ms/GeneralAI}{https://aka.ms/GeneralAI}}
}

\date{}

\begin{document}
\maketitle

\vspace{-0.2in}
\begin{abstract}

This technical report presents \textbf{\our{}}, a vision Transformer that can process gigapixel images in an end-to-end manner.
Specifically, we split the gigapixel image into a sequence of millions of patches and project them linearly into embeddings.
\longnet{}~\cite{longnet} is then employed to model the extremely long sequence, generating representations that capture both short-range and long-range dependencies.
The linear computation complexity of \longnet{}, along with its distributed algorithm, enables us to overcome the constraints of both computation and memory.
We apply \our{} in the field of computational pathology, aiming for cancer diagnosis and prognosis within gigapixel whole-slide images.
Experimental results demonstrate that \our{} effectively encodes gigapixel images and outperforms previous state-of-the-art methods on cancer subtyping and survival prediction.
Code and models will be available at \url{https://aka.ms/LongViT}.

\end{abstract}

\vspace{-0.2in}
\begin{figure*}[ht]
\centering
\includegraphics[width=0.9\columnwidth]{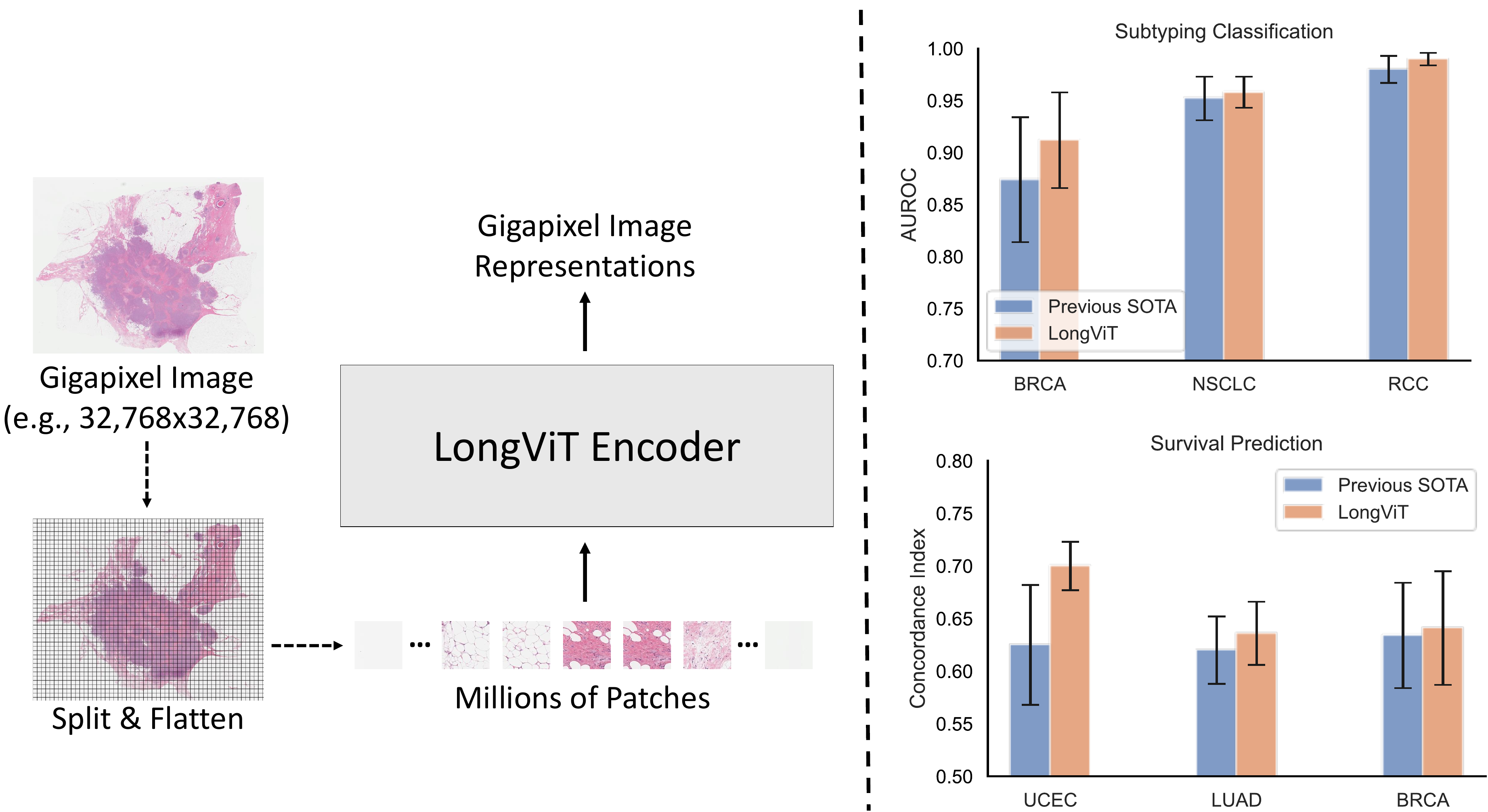}
\caption{
\our{} can effectively process gigapixel images in an end-to-end manner via \longnet{}~\cite{longnet}.
We evaluate \our{} in the field of computational pathology.
\our{} outperforms previous SOTA methods across six subtyping classification and survival prediction tasks.
}
\label{fig:lingvit-summary}
\end{figure*}

\section{Introduction}
\label{sec:intro}

Vision Transformers~\cite{vit,deit,beit} have achieved promising performance across a wide range of vision tasks.
However, there are still significant challenges in applying vision Transformers to process gigapixel images, such as whole slide images (WSIs) in computational pathology~\cite{hipt}, due to the computation and memory requirements brought about by the enormous number of pixels.

\begin{figure*}[t]
\centering
\includegraphics[width=1.0\columnwidth]{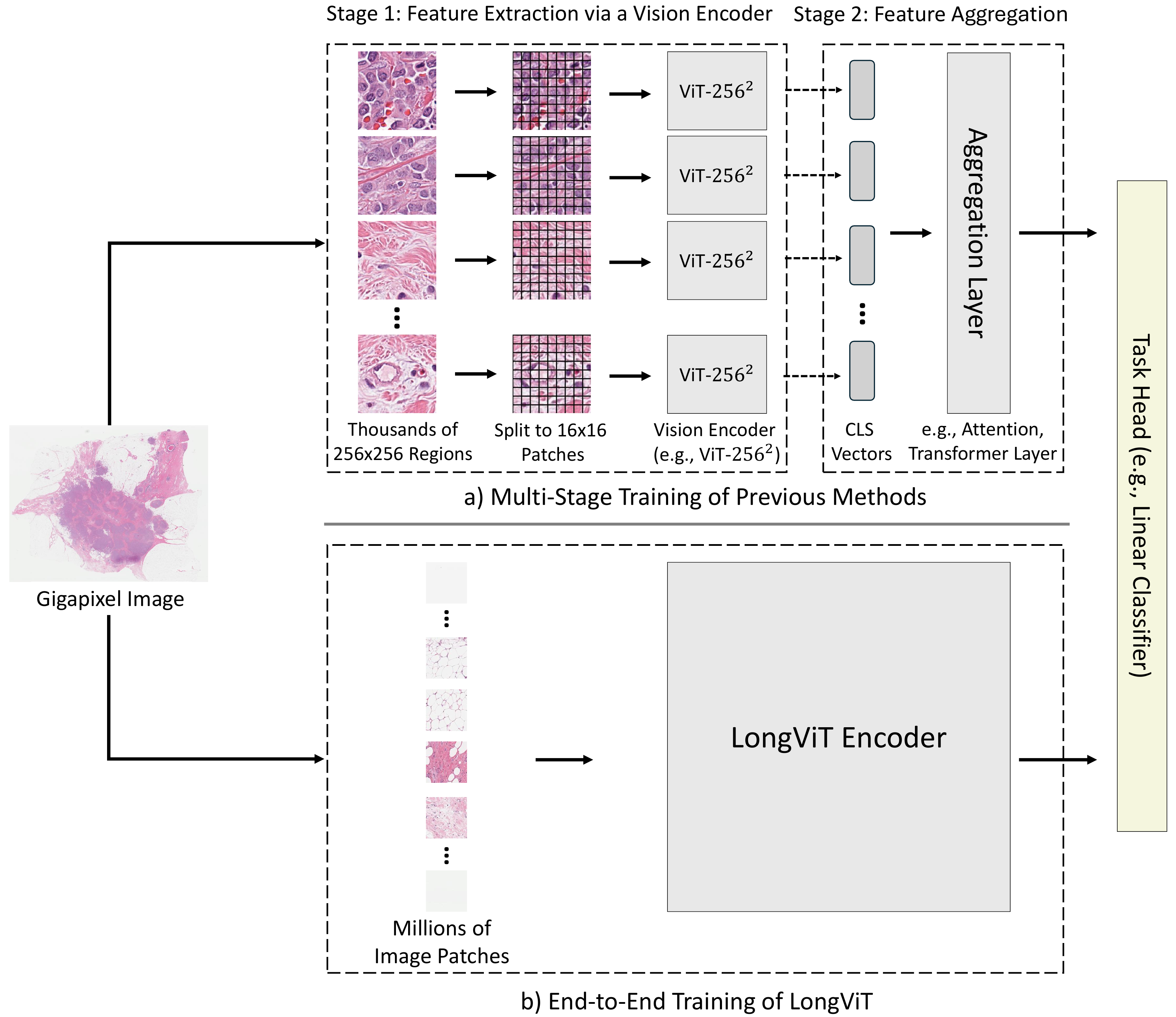}
\caption{
Overview of \our{} and previous multi-stage methods~\cite{mil,calm,transmil,hipt}.
We split a gigapixel image into a sequence of millions of patches, perform linear projection, and add positional embeddings.
Next, we use \longnet{}~\cite{longnet} to model the very long sequence to obtain representations of the whole gigapixel image.
Different from previous methods using multi-stage training, \our{} learns representations of gigapixel images in a simple end-to-end manner.
ViT-$256^{2}$ indicates pretrained vision Transformer working on a 256$\times$256 resolution.
The region size, patch size and vision encoder used in previous works might be different from the example.
}
\label{fig:lingvit}
\end{figure*}

In this technical report, we introduce \our{}, a vision Transformer that can efficiently encode gigapixel images.
We first reshape a gigapixel image into millions of patches and perform linear projection to obtain the patch embeddings.
We then use \longnet{}~\cite{longnet} to encode the extremely long sequence of patch embeddings, learning the representations of the gigapixel image.
\longnet{}~\cite{longnet} is a Transformer~\cite{transformer} variant that can model very long sequences via dilated attention.
Its linear computation complexity and distributed algorithm allow the model to handle gigapixel images in an efficient way.

We apply \our{} in the field of computational pathology, which aims to assist in the diagnosis and prognosis of cancers through the analysis of gigapixel whole slide images.
As shown in Figure~\ref{fig:lingvit}, previous works~\cite{mil,ds-mil,calm,mcat,hipt} mostly adopt a multi-stage framework. 
They first segment the tissue region of a slide image and divide it into smaller regions (usually, 256$\times$256 pixels).
A pretrained vision encoder~\cite{dino} is employed for feature extraction of these smaller regions.
These features are then fed into an aggregation layer, such as attention-based pooling, Transformer, and hierarchical Transformer to obtain the representations of whole slide images.
Different from previous works, \our{} learns representations of gigapixel images in an end-to-end manner, simplifying previous multi-stage encoding into a single stage.
The model directly takes the whole slide image as input, without the need for data preprocessing such as tissue segmentation.
In addition, \our{} effectively captures both short-range and long-range dependencies from millions of patches to fit different tasks in computational pathology.

Following previous work~\cite{hipt}, we perform self-supervised pretraining on about 10k diagnostic slides from The Genome Cancer Atlas (TCGA~\cite{tcga}) using DINO~\cite{dino} as the pretraining objective.
The model is then finetuned on cancer subtyping classification and survival prediction tasks.
Experimental results demonstrate that \our{} achieves remarkable performance across six subtyping classification and survival prediction tasks, outperforming various state-of-the-art approaches.
Ablation studies of different input image resolutions further show the effectiveness of \our{} to encode extremely large images.

\section{Experiments}
\label{sec:exps}

Given a gigapixel image, we first divide the image into a sequence of millions of fixed-size patches.
We then generate patch embeddings via linear projection and add learnable 1D position embeddings.
Next, we employ \longnet{} to model the very long sequence to obtain patch representations that contain contextual information.
We use average pooling on the patch representations to generate the representation of the whole gigapixel image.
Due to the computation and memory constraints, we split the sequence of millions of patches along the sequence dimension and use the distributed algorithm of \longnet{} to parallelize the training across multiple GPU devices.

We apply \our{} on cancer subtyping classification and survival prediction in computational pathology.
Following HIPT~\cite{hipt}, we pretrain a model of the same size as ViT-S on about 10,000 diagnostic whole slide images from more than 30 different types of cancer in TCGA using DINO~\cite{dino} framework.
To reduce the computation cost, we perform pretraining at a resolution of 1,024.
We randomly crop 100 regions from each whole slide image at a 20$\times$ objective, with dimensions ranging from 1,024 to 1,536 pixels in both width and height.
The patch size is $32\times32$.
Detailed hyperparameters are present in Table~\ref{tbl:hyperparam:pretraining} in the Appendix.
For finetuning, we resize the whole slide image to a very large size (e.g., $32,768\times32,768$), and then feed the resized image to \our{} to obtain its representations.
We perform 2D interpolation of the pretrained position embeddings as in standard vision Transformers~\cite{vit}.
We perform pretraining on 64 V100 GPUs and finetuning on 8 V100 GPUs.

\begin{table}[t]
\centering
\begin{tabular}{l|c|c|c}
\toprule
\bf Model & \bf BRCA Subtyping & \bf NSCLC Subtyping & \bf RCC Subtyping \\ \midrule
MIL~\cite{calm} & $0.778 \pm 0.091$ & $0.892 \pm 0.042$ & $0.959 \pm 0.015$ \\
CLAM-SB~\cite{calm} & $0.858 \pm 0.067$ & $0.928 \pm 0.021$ & $0.973 \pm 0.017$ \\
DeepAttnMISL~\cite{deepattnmisl} & $0.784 \pm 0.061$ & $0.778 \pm 0.045$ & $0.943 \pm 0.016$ \\
GCN-MIL~\cite{gcn-mil} & $0.840 \pm 0.073$ & $0.831 \pm 0.034$ & $0.957 \pm 0.012$ \\
DS-MIL~\cite{ds-mil} & $0.838 \pm 0.074$ & $0.920 \pm 0.024$ & $0.971 \pm 0.016$ \\
HIPT~\cite{hipt} & $0.874 \pm 0.060$ & $0.952 \pm 0.021$ & $0.980 \pm 0.013$ \\
\midrule
\bf \our & $\mathbf{0.912 \pm 0.046}$ & $\mathbf{0.958 \pm 0.015}$ & $\mathbf{0.990 \pm 0.006}$ \\
\bottomrule
\end{tabular}
\vspace{0.2cm}
\caption{Results of subtyping classification tasks. Following HIPT~\cite{hipt}, we report 10-fold cross-validated AUC performance. 
The macro-averaged AUC performance across the three subtypes is reported for RCC subtyping.
Results of HIPT and other models are from \citep{hipt}.
For BRCA and RCC subtyping, the input resolution is $32,768\times32,768$ and the length of the resulting sequence is $1,048,576$.
For NSCLC subtyping, the input resolution is $16,384\times16,384$ and the length of the sequence is $262,144$.}
\label{tbl:exps:subtyping}
\end{table}

\begin{table}[t]
\centering
\begin{tabular}{lccc}
\toprule
\bf Model & \bf UCEC & \bf LUAD & \bf BRCA \\ \midrule
\multicolumn{4}{l}{\emph{w/ genomic features}} \\
\quad Deep Sets~\cite{deepsets} & $0.598 \pm 0.077$ & $0.616 \pm 0.027$ & $0.521 \pm 0.079$ \\
\quad Attention MIL~\cite{mil} & $0.614 \pm 0.052$ & $0.563 \pm 0.050$ & $0.551 \pm 0.077$ \\
\quad DeepAttnMISL~\cite{deepattnmisl} & $0.615 \pm 0.020$ & $0.595 \pm 0.061$ & $0.545 \pm 0.071$ \\
\quad MCAT~\cite{mcat} & $0.622 \pm 0.019$ & $0.620 \pm 0.032$ & $0.580 \pm 0.069$ \\
\midrule
\multicolumn{4}{l}{\emph{w/o genomic features}} \\
\quad HIPT~\cite{hipt} & - & $0.538 \pm 0.044$ & $0.634 \pm 0.050$ \\
\quad Deep Sets~\cite{deepsets} & $0.500 \pm 0.000$ & $0.496 \pm 0.008$ & $0.500 \pm 0.000$ \\
\quad Attention MIL~\cite{mil} & $0.625 \pm 0.057$ & $0.559 \pm 0.060$ & $0.564 \pm 0.050$ \\
\quad DeepAttnMISL~\cite{deepattnmisl} & $0.597 \pm 0.059$ & $0.548 \pm 0.050$ & $0.524 \pm 0.043$ \\
\quad \bf \our & $\mathbf{0.700 \pm 0.023}$ & $\mathbf{0.636 \pm 0.030}$ & $\mathbf{0.641 \pm 0.054}$ \\
\bottomrule
\end{tabular}
\vspace{0.2cm}
\caption{Results of survival prediction across three cancer datasets. 
Following MCAT~\cite{mcat}, we report 5-fold cross-validated concordance index (c-Index) performance.
Results of HIPT are from \citep{hipt}. Results of MCAT and other models are from \citep{mcat}.
The input resolution is $1,024\times1,024$ ($1,024$ patches) for UCEC dataset, $32,768\times32,768$ ($1,048,576$ patches) for LUAD dataset, and $16,384\times16,384$ ($262,144$ patches) for BRCA dataset.
}
\label{tbl:exps:survival}
\end{table}

\subsection{Cancer Subtyping}
\label{sec:subtyping}

The task aims to classify the cancer subtypes based on whole slide images.
Following previous works~\cite{calm,hipt}, we evaluate the model on three multi-class subtyping problems including Invasive Breast
Carcinoma (BRCA) subtyping, Non-Small Cell Lung Carcinoma (NSCLC) subtyping, and Renal Cell Carcinoma (RCC) subtyping.
BRCA and NSCLC subtyping are binary classification problems, while RCC subtyping is a ternary classification problem.
We conduct experiments using 10-fold cross-validation and the same setting as HIPT~\cite{hipt}.  
Cross-validated AUC is reported.
The model is trained using cross-entropy loss.
As present in Table~\ref{tbl:exps:subtyping}, \our{} achieves promising performance and outperforms previous approaches.
Our model achieves a 3.8\% improvement in BRCA subtyping, 1.0\% improvement in RCC subtyping, and 0.6\% improvement in NSCLC subtyping.
We train the model for 10 epochs with a batch size of 8 and a learning rate of 5e-5.
For the experiments using $32,768\times32,768$ input resolution, we parallelize the training and use a batch size of 2 with 4 gradient accumulation steps.
More detailed finetuning hyperparameters can be found in Table~\ref{tbl:hyperparam:finetuning} in Appendix.

\begin{table}[t]
\centering
\small
\begin{tabular}{l|c|c|c}
\toprule
\textbf{Image Resolution (\#Patches)} & \bf BRCA Subtyping & \bf NSCLC Subtyping & \bf RCC Subtyping \\ \midrule
$1,024\times1,024$ ($1,024$) & $0.869 \pm 0.055$ & $0.916 \pm 0.017$ & $0.966 \pm 0.022$ \\
$4,096\times4,096$ ($16,384$) & $0.893 \pm 0.027$ & $0.938 \pm 0.012$ & $0.983 \pm 0.011$ \\
$8,192\times8,192$ ($65,536$) & $0.907 \pm 0.052$ & $0.952 \pm 0.019$ & $0.985 \pm 0.007$ \\
$16,384\times16,384$ ($262,144$) & $0.906 \pm 0.041$ & $0.958 \pm 0.015$ & $0.989 \pm 0.006$ \\
$32,768\times32,768$ ($1,048,576$) & $0.912 \pm 0.046$ & $0.955 \pm 0.017$ & $0.990 \pm 0.006$ \\
\bottomrule
\end{tabular}
\vspace{0.2cm}
\caption{Ablation study on the impact of image resolution in subtyping classification.}
\label{tbl:exps:ablation:subtyping_image_resolution}
\end{table}

\begin{table}[t]
\centering
\small
\begin{tabular}{l|c|c|c}
\toprule
\textbf{Image Resolution (\#Patches)} & \bf UCEC & \bf LUAD & \bf BRCA \\ \midrule
$1,024\times1,024$ ($1,024$) & $0.700 \pm 0.023$ & $0.573 \pm 0.016$ & $0.643 \pm 0.051$ \\
$4,096\times4,096$ ($16,384$) & $0.658 \pm 0.060$ & $0.598 \pm 0.018$ & $0.594 \pm 0.046$ \\
$8,192\times8,192$ ($65,536$) & $0.671 \pm 0.056$ & $0.604 \pm 0.031$ & $0.638 \pm 0.040$ \\
$16,384\times16,384$ ($262,144$) & $0.643 \pm 0.055$ & $0.627 \pm 0.038$ & $0.641 \pm 0.054$ \\
$32,768\times32,768$ ($1,048,576$) & $0.659 \pm 0.035$ & $0.636 \pm 0.030$ & $0.589 \pm 0.065$ \\
\bottomrule
\end{tabular}
\vspace{0.2cm}
\caption{Ablation study on the impact of image resolution in survival prediction.}
\label{tbl:exps:ablation:survival_image_resolution}
\end{table}

\subsection{Survival Prediction}

The task aims to predict the relative risk of cancer death.
We evaluate the model on three cancer datasets from TCGA: Uterine Corpus Endometrial Carcinoma (UCEC), Lung Adenocarcinoma (LUAD), and Breast Invasive Carcinoma (BRCA).
We perform finetuning in a 5-fold cross-validation using the same setting and loss function as MCAT~\cite{mcat}.
We use the same hyperparameters as subtyping tasks (Section~\ref{sec:subtyping}) to train the model.
We report the cross-validated concordance index (c-Index) of \our{} and previous models in Table~\ref{tbl:exps:survival}.
Our model also achieves superior performance across three datasets and outperforms previous state-of-the-art models without using genomic features.

\subsection{Ablation Study}

We conduct ablation studies using images of different resolutions during finetuning.
The results on subtyping classification and survival prediction are present in Table~\ref{tbl:exps:ablation:subtyping_image_resolution} and Table~\ref{tbl:exps:ablation:survival_image_resolution}.
We use the same hyperparameters except the segment length used in \longnet{} encoder, which varies with the input resolution.
For most tasks, including subtyping on three datasets and LUAD survival prediction, the results improve as the resolution of the input image increases.
Using a large resolution is essential for these tasks.
Experimental results also demonstrate that \our{} effectively learns representations of very large images.
We will also try finetuning on larger images in future work.
However, for UCEC and BRCA survival prediction, especially for UCEC, we find a different trend.
This observation could potentially be attributed to the task data or the $1,024$ input image resolution utilized during pretraining.
We will explore these possible underlying factors in the future work.

\section{Conclusion}
\label{sec:conclusion}

This technical report introduces \our{}, which aims to scale vision Transformers to gigapixel images in an end-to-end manner.
We use \longnet{} to encode the long sequence of millions of patches, which is obtained by directly splitting the gigapixel image, to learn representations of the whole image.
\our{} can efficiently process extremely large images and capture both short-range and long-range dependencies.
We conduct experiments on cancer subtyping classification and survival prediction tasks in computational pathology.
\our{} achieves promising performance and outperforms various state-of-the-art models.

\bibliographystyle{alpha}
\bibliography{longvit}

\newpage
\appendix

\begin{table}[H]
\centering
\small
\begin{tabular}{l|c}
\toprule
\bf Hyperparameters & \bf \our{} \\
\midrule
Layers & 12 \\
Hidden size & 384 \\
FFN inner hidden size & 1,536 \\
Attention heads & 16 \\
Patch size & $32 \times 32$ \\
\midrule
Training epochs & 100 \\
Batch size & 1,024 \\
Peak learning rate & 2e-3 \\
Warmup epochs & 10 \\
Drop path & 0.1 \\
Input resolution & $1,024^2$ \\
Segment lengths & \{64, 128, 256, 512, 1024\} \\
Dilated ratios & \{1, 2, 4, 8, 16\} \\
\midrule
Global crops scale & (0.4, 1.0) \\
Global crops size & $1,024^2$ \\
Local crops scale & (0.05, 0.4) \\
Local crops size & $512^2$ \\
\bottomrule
\end{tabular}
\vspace{2mm}
\caption{
Hyperparameters for pretraining \our{}.
}
\label{tbl:hyperparam:pretraining}
\end{table}

\begin{table}[ht]
\centering
\small
\begin{tabular}{lc}
\toprule
\textbf{Hyperparameters} & \textbf{\our{}} \\ 
\midrule
Peak learning rate & 5e-5 \\
Fine-tuning epochs & 10 \\
Warmup epochs & 1 \\
Batch size & 8 \\
Weight decay & 0.05 \\
Drop path & 0.1 \\
Dilated ratios & \{1, 2, 4, 8, 16\} \\
\bottomrule
\end{tabular}
\vspace{0.2cm}
\caption{Finetuning hyperparameters of \our{}}
\label{tbl:hyperparam:finetuning}
\end{table}

\begin{table}[ht]
\centering
\small
\begin{tabular}{lc}
\toprule
\textbf{Input Resolution} & \textbf{Segment Lengths} \\ 
\midrule
$1,024^2$ & \{64, 128, 256, 512, 1024\} \\
$4,096^2$ & \{1024, 2048, 4096, 8192, 16384\} \\
$8,192^2$ & \{1024, 4096, 8192, 16384, 65536\} \\
$16,384^2$ & \{1024, 4096, 16384, 65536, 262144\} \\
$32,768^2$ & \{1024, 4096, 32768, 262144, 1,048576\} \\
\bottomrule
\end{tabular}
\vspace{0.2cm}
\caption{Hyperparameters of segment lengths for different input resolutions.}
\label{tbl:hyperparam:segment_lengths}
\end{table}

\end{document}